# Memory Based Machine Intelligence Techniques in VLSI hardware

Alex Pappachen James,
Machine Intelligence Research Group,
Indian Institute of Information Technology and Management-Kerala
Email: apj@ieee.org

Abstract: *We briefly introduce the memory based approaches to emulate machine intelligence in VLSI hardware, describing the challenges and advantages. Implementation of artificial intelligence techniques in VLSI hardware is a practical and difficult problem. Deep architectures, hierarchical temporal memories and memory networks are some of the contemporary approaches in this area of research. The techniques attempt to emulate low level intelligence tasks and aim at providing scalable solutions to high level intelligence problems such as sparse coding and contextual processing.*

Neocortex in human brain that accounts for 76% of the brain's volume can be seen as a control unit that is involved in the processing of intelligence functions such as sensory perception, generation of motor commands, spatial reasoning, conscious thought and language. The realisation of neocortex like functions through algorithms and hardware implementation has been the inspiration to majority of state of the art artificial intelligence systems and theory.

Intelligent functions are often required to be performed in environments having high levels of natural variability, which makes individual functions of the brain highly complex. On the other hand, the huge number of neurons in neocortex and their interconnections with each other makes the brain structurally very complex. As a result, the mimicking of neocortex is exceptionally challenging and difficult.

Human brain is highly modular and hierarchical in structure and functionality. Each module in a human brain has evolved in structure and is trained to deal with very specific task. The interaction between the modules results in the ability of human brain to understand and respond to complex cognitive task. The interaction of the module follows a hierarchical organization, where sensory processes form the bottom part of hierarchy and decision making processes form the top most part of the hierarchy.

Even with the challenges of complexity, since the early 1950's researchers have tried to model the neuron and the cortex through mathematical, biological, algorithmic, bio-memetic and neuromporphic circuits. Computational, cognitive, and memory capability of human brain has inspired researchers to emulate brain function or capability in artificial intelligent systems [1-6]. The artificial intelligent systems are implemented in software or hardware and often derived from mathematical or logical modelling of brain capability, examples of which are artificial neural networks, fuzzy logic, evolutionary computation, Bayesian networks, expert systems, case based reasoning, and behaviour based artificial intelligence. They find application in knowledge based problems, pattern recognition problems, optimization problems and adaptive control problems. Dedicated artificial intelligent hardware chips have many advantages over software implementations, but the existing hardware implementations are limited to low functional complexity and small number of inputs. As an example, the architecture of artificial neural networks needs many neural nodes and weights that require many amplifiers and memory elements. Large area is needed to place these components, and with an increase in the number of neural nodes, the number of interconnections between the neural nodes in different layers becomes high. This results in impossible cell routing in a limited chip area, while large area requirement of the amplifiers and memory elements limit the size of the network. Similarly, fuzzy logic chip implementation with large number of inputs is also not possible due to hardware complexity.

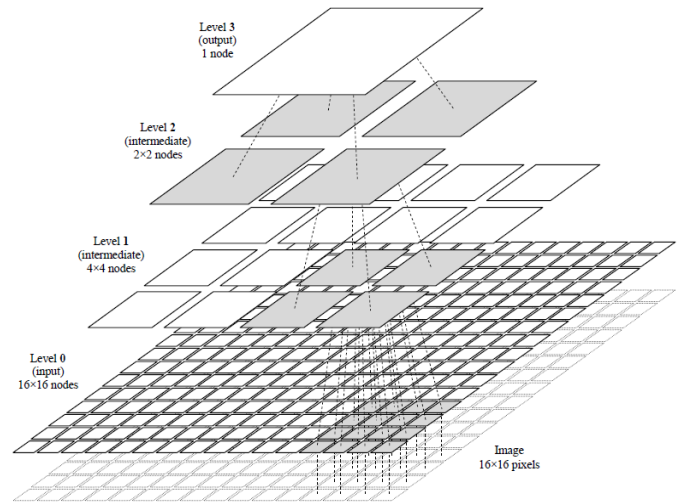

Fig 1: The hierarchical nature of the hierarchical temporal memory networks

The idea of human intelligence cannot exist without understanding the concept of memory. Human memory has continuously gained attention of neuroscientists over the last 50 years, and only very recently has captured the interest of machine intelligence researchers. In so-called intelligent machines that we know today, the concept of memory is essential to the idea of implementing intelligent computing, without which modelling any event or simulating any learning mechanism becomes almost impossible. Although it becomes very trivial that memory is essential, the idea of memory being the basis of intelligence is less understood and investigated. Most modern computational intelligence tries to emulate the intelligence using memory in one way or the other. Rather than following an algorithmic approach to computational intelligence, it becomes more apparent by looking at hierarchical nature of human brain and network of neurons, that human memories and cognitive mechanisms that are organized in a hierarchical manner form network of neurons can be modelled and emulated to form new techniques for computational intelligence.

Perhaps one of the first approaches in network research is the so-called artificial neural networks, which focused on solving pattern recognition and optimization problems by training the network weights through supervised or semi-supervised approach. Multistage Hubel-Wiesel architectures (MHWA) is a deep architecture that consists of alternating layers of feature detectors and local pooling of features. Some of the implementations of MHWA are neocognitron, convolution networks, HMAX and its variants. However, the practical implementation of the artificial neural networks in hardware is not an easy task and in many cases is an impossible task.

With increase in number of artificial cells (or nodes) the number of cross-over wiring required to implement the artificial neural network increases exponentially, which means increased design complexity and reduced fault tolerance with the individual network cells. Smaller the semiconductor devices, bigger the problems associated with cross-over wiring and designing trainable weights. The weights are often implemented as controlled memory devices, where the weights are accessed and set through conventional addressing logic. Although this approach makes it easy to train, it is perhaps not the most optimal way to implement the hardware weights. A more realistic implementation that matches closely with neural network model in hardware would require an integrated approach to hardware implementation, where the individual cells are self-contained with its own dedicated weights. However this is not a trivial task especially when implementing large scale artificial neural networks, as addressing memories within each cell and collaborating with the network cells while learning require addition control circuits increasing the complexity of design and implementation.

A recent extension to the deep architecture approach is the hierarchical temporal networks (HTM) shown in Fig 1 that use the deep architectures in a hierarchical manner. A high level of modularity and hierarchy in function and structure is a key feature of neocortex. HTM uses several of the existing concepts in pattern recognition to form a networks approach to prediction of objects and sequences across time.

Memory networks [6] overcome such implementation issues that occur with cross-over wiring and introduce a high level of hierarchical modularity which is unseen in conventional artificial neural networks. Recent results show that hierarchical networks such as that inspired from cortical learning algorithms (eg. HTM) perform well against changes in untrained natural variability. Some of the small scale and generalized hardware implementations have produced promising results and insights into the area of hierarchical memory networks.

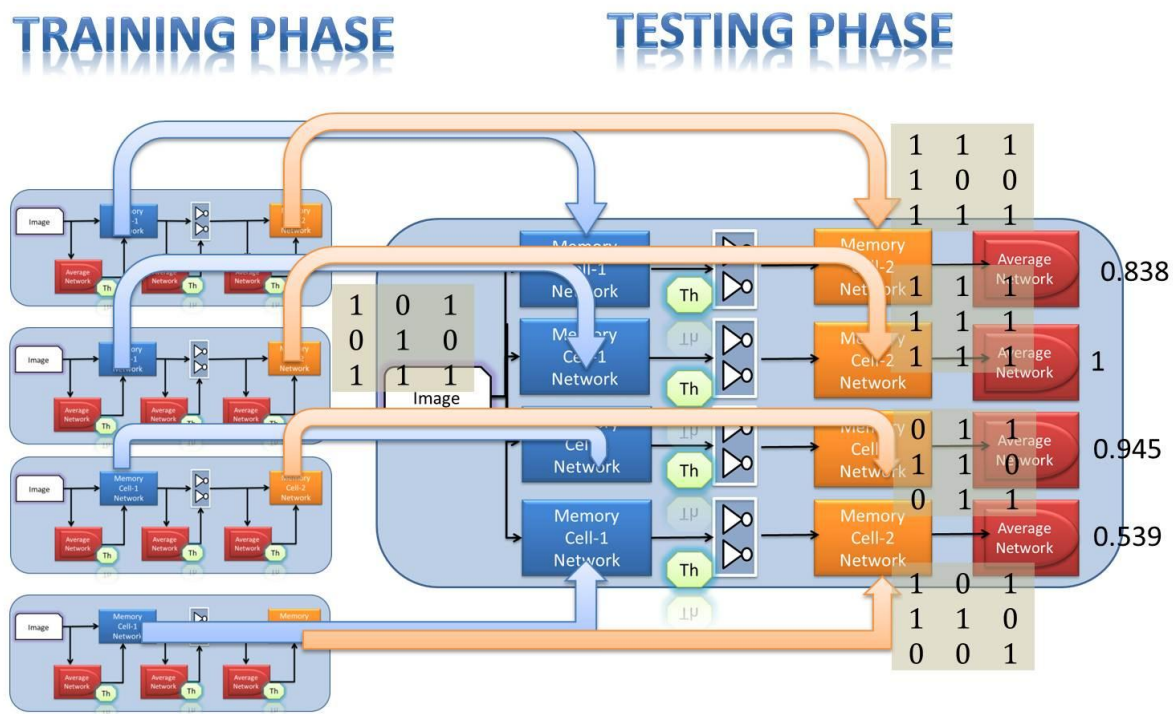

Fig 2: The block level illustration of the memory network in a multiclass high-dimensional pattern recognition problem. Each memory cell network corresponds to that reported in [6].

The cognitive memory networks on the other hand use a far simplistic approach in implementation and learning as reported in [6]. An extended version of the memory network reported in [6] is shown in Fig 2, where it is applied to a practical face recognition problem in real-time environments. It follows a supervised learning approach using various stages of genetic learning for fixing the cell weights. The weights in a cognitive memory network are equivalent resistors of 2 or 3 terminal resistive memory devices. Since equivalent resistors of the memory elements are used for training the network, the actual use of memory elements as storage become implicative than formal. The network as a whole through its various layers memorizes an event, which means it allows for redundancy of individual cells, making the network fault tolerant and robust to changes in natural variability. Further, scaling such a network to the extent that a typical human memory would like to go becomes almost a trivial task, which is typically not possible with conventional artificial neural network type implementations. These properties of the memory networks make it a promising alternative to understand and emulate intelligence in machines. In addition, the hierarchical nature of the network can be explored to implement image processing and data compression application, such as done with autocorrelation neural networks.

The potential long term implication of memory network architectures is that it can perform sensory detection, perception, cognition, memory, and attention functions. However maximum information storage time of the network depends on the maximum retention time of memory used. In terms of hardware implementation this needs memories that can do large number of cycles, less area and long data retention time which may be made possible with QsRAM.

The modularity of the network can be put use to create large structures in hardware, which increases the ability of the network to handle more complex functions and a way to create digital intelligence. The ability of the network to handle complex tasks depends on the physical complexity of the final architecture, and effectively indicates to the arrangement of memory elements, layers, and number of cell inputs, which can bring about various network structures. The hardware memory network can have structures that can be better or worse for a problem it can handle. Moreover, the selection of the network structure depends on the functional complexity and nature of task. In the examples we have provided what a simple cell type can do when we replicate it throughout the network. The number of inputs that can be connected to a cell is restricted to a maximum of 6 due to the hardware issues associated with the inverter. Further the nonlinear inverter function in the cell can introduce errors in its output, which can accumulate from a layer to next layer. But by simulation experiments it can be shown that these errors are marginal and it does not affect the overall working of the network. The network is stable to the tolerance or sensitivity of resistance variations, which implies that small deviations on the trained values do not affect the network functionality.

Memory network implementation for a 1296 inputs and 12 bit output network in hardware needs 1812 bits of memory and 264 inverters, with a input resolution of 1/1296 with 3 layers of cells [6].So if we extend the memory to 1 gigabit with 12 bit output implies a input resolution of 1/766958445 with 10 layers of cells, and this network will be able to solve any other complex issues related with the examples addressed. The memory network chip with $10^6$ gigabits could perform most of the tasks that a human brain does, while a working memory of 1 or 2 gigabits should be enough to compete with most cognitive abilities of brain. The current technologies provide memory area that can compete with the biological memories and is under-utilized due to technological constraints. With more technologies under development large semiconductor memories will be made possible in coming years. This will help build large memory network chips which can match the cognitive capability of human brain, and can be the starting point to a fully functional autonomous artificial intelligence.

.